\documentclass[11pt]{article}

\usepackage[preprint]{latex/acl}

\usepackage{times}
\usepackage{latexsym}

\usepackage[T1]{fontenc}

\usepackage[utf8]{inputenc}

\usepackage{microtype}

\usepackage{inconsolata}

\usepackage{graphicx}
\usepackage{booktabs} 
\usepackage{amsmath, amssymb}
\usepackage{float}  
\usepackage{xcolor}  
\definecolor{MyHexPurple}{HTML}{8E7DBE}
\definecolor{MyHexBlue}{HTML}{A6D6D6}
\usepackage[most]{tcolorbox}
\usepackage{xcolor}
\usepackage{subcaption}

\newtcolorbox{userinputbox}{
  enhanced,
  colback=gray!15,
  colframe=gray!60,
  title=User Input,
  fonttitle=\ttfamily\bfseries\footnotesize\color{black},
  fontupper=\ttfamily\footnotesize,
  arc=3mm,
  sharp corners=south,
  boxrule=0.6pt,
  top=2mm,bottom=2mm,left=3mm,right=3mm,
  attach boxed title to top left={
    xshift=2mm,
    yshift*=-2mm
  },
  boxed title style={
    colback=gray!25,
    colframe=gray!60,
    arc=2mm
  },
  before skip=0pt,
  after skip=0pt
}

\newtcolorbox{assistantprefixbox}{
  enhanced,
  colback=blue!10,          
  colframe=gray!60,
  title=Assistant Prefix,
  fonttitle=\ttfamily\bfseries\footnotesize\color{black},
  fontupper=\ttfamily\footnotesize,
  arc=3mm,
  sharp corners=north,
  boxrule=0.6pt,
  top=2mm,bottom=2mm,left=3mm,right=3mm,
  attach boxed title to top left={
    xshift=2mm,
    yshift*=-2mm
  },
  boxed title style={
    colback=blue!20,
    colframe=gray!60,
    arc=2mm
  },
  before skip=0pt
}

\title{When English Isn’t the Best Teacher: Source Language Effects in Cross-Lingual In-Context Learning}

\author{
 \textbf{Fred Philippy\textsuperscript{1}},
 \textbf{Siwen Guo\textsuperscript{2}},
 \textbf{Jacques Klein\textsuperscript{1}},
 \textbf{Tegawendé F. Bissyandé\textsuperscript{1}}
\\
\\
 \textsuperscript{1}Snt, University of Luxembourg, Luxembourg
 \\
 \textsuperscript{2}Luxembourg Institute of Science and Technology, Luxembourg
\\
 \small{
   \textbf{Correspondence:} \href{mailto:fred.philippy@uni.lu}{fred.philippy@uni.lu}
 }
}

\begin{document}
\maketitle

 \begin{abstract}
Cross-lingual transfer in multilingual NLP has been widely explored in supervised fine-tuning contexts, where factors like data availability and linguistic similarity largely determine transfer quality. As the field shifts toward few-shot In-Context Learning (ICL), it is often presumed that insights from fine-tuning carry over unchanged. Yet this assumption has not been rigorously evaluated, leaving open the question of how to choose source languages for cross-lingual ICL. We conduct a broad empirical study of cross-lingual transfer in ICL spanning seven tasks, six models, and a typologically diverse set of languages. We further analyze language confusion, a key obstacle for generative tasks in cross-lingual ICL. Our results show that conventional fine-tuning-based expectations do not consistently apply in the ICL regime and point to alternative heuristics for selecting source languages effectively.
\end{abstract}
\section{Introduction}

Large language models (LLMs) have dramatically improved performance across a wide range of NLP tasks, yet their capabilities remain uneven across the world’s languages. A core reason is the stark imbalance in training data availability: high-resource languages such as English dominate pretraining corpora, while many others remain severely underrepresented. Much of the recent work on multilingual LLMs therefore focuses on cross-lingual transfer, where knowledge gained from high-resource languages is leveraged to improve performance in low-resource ones. Prior research, typically using supervised fine-tuning setups, has shown that cross-lingual transfer is far from uniform: factors such as linguistic similarity, lexical overlap, and writing system strongly influence how well knowledge transfers from a source language to a target one \citep{pires-etal-2019-multilingual, KWMR20, muller-etal-2021-first, philippy-etal-2023-towards}.
However, these findings largely reflect the era of task-specific fine-tuning, where models are explicitly trained on labeled data in a source language and then evaluated in a different target language. With the rise of large, instruction-tuned LLMs, the field is increasingly shifting toward In-Context Learning (ICL) approaches that avoid parameter updates altogether \citep{brown2020fewshot}. ICL departs from traditional fine-tuning by allowing models to perform tasks solely through examples and instructions provided at inference time. Instead of relying on parameter modification, LLMs condition on the prompt itself and learn ``in context'' from natural language demonstrations, task descriptions, or other structured cues. This paradigm has proven surprisingly effective across diverse tasks, and recent work shows that it can also support cross-lingual transfer: models can use demonstrations in one language to perform a task in another \citep{winata-etal-2022-cross, tu-etal-2025-blessing}.
Yet, unlike in the fine-tuning setting, it remains unclear which source languages are most effective for ICL for a given target language, and whether the factors known to influence fine-tuning transfer (e.g., linguistic similarity) apply equally in this new paradigm.

In this study, we investigate how source–target language relationships shape cross-lingual transfer in ICL. By systematically examining the effects of typology, writing systems, embedding-based alignment, and resource availability, our goal is to provide clearer guidance on how to select effective source languages for ICL, especially when working with low-resource target languages.

Our analysis reveals that assumptions inherited from fine-tuning do not always hold in ICL. In contrast to established findings in fine-tuning-based cross-lingual transfer, we observe that the target language itself is its most effective source language in only about 24\% of cases, and that English, despite its disproportionate presence in pretraining corpora, emerges as the worst source language in about 16\% of our experiments. Moreover, we show that linguistic similarity, which is widely regarded as the strongest predictor of cross-lingual transfer in supervised fine-tuning, plays a far less important role in cross-lingual ICL.
In additional experiments focused on language confusion, a phenomenon known to hinder cross-lingual transfer in generative tasks, we likewise uncover substantial discrepancies between source and target languages during ICL.

Altogether, our findings not only call for rethinking cross-lingual transfer in ICL but also point to a promising direction: the least supported languages in LLMs, particularly low-resource non-Latin-script languages, may in fact serve as unexpectedly strong sources from which other languages can benefit.

\begin{table*}[!ht]
\centering
\begin{tabular}{llllrrrc}
\toprule
Paper & Dataset & Task & \(n\) & $|\mathcal{L}|$ & $|\mathcal{L}^*|$ & Metric \\
\midrule
\citet{yang-etal-2019-paws} & PAWS-X &
Paraphrase Identification & 2k & 7 & 7 & F1 \\

\citet{lin-etal-2022-shot} & XStoryCloze &
Commonsense Reasoning & 1.51k & 11 & 9 & F1 \\
\citet{singh-etal-2025-global} & Global-MMLU & 
Knowledge Comprehension & 14k & 42 & 17 & Acc. \\

\citet{conneau-etal-2018-xnli} & XNLI &
Natural Language Inference & 5.01k & 15 & 12 & F1 \\

\citet{shi2022languagemodelsmultilingualchainofthought} & MGSM &
Mathematical Reasoning & 250 & 11 & 11 & Acc. \\

\citet{adelani-etal-2024-sib} & SIB-200 &
Topic Classification & 204 & 206 & 18 & F1 \\

\citet{ponti-etal-2020-xcopa} & XCOPA &
Commonsense Reasoning & 500 & 11 & 7 & F1 \\
\bottomrule
\end{tabular}
\caption{
Multilingual datasets included in our study. 
Here, \(n\) denotes the number of test samples per language, 
\(|\mathcal{L}|\) the full language coverage of each dataset, 
and \textbf{\(|\mathcal{L}^*|\)} the subset of languages we use as source and target languages.
}
\label{tab:multilingual_datasets}
\end{table*}

\section{Related Work}
Prior research indicates that demonstration language selection and configuration significantly impact cross-lingual ICL performance. Early studies found that cross-lingual prompting often surpasses monolingual baselines, though linguistic proximity is not always a reliable predictor of success \citep{winata-etal-2022-cross}. Recent work explores diverse prompt structures, including mixed-language demonstrations \citep{kim2024crosslingualqakeyunlocking} and code-switching transitions to English \citep{yoo2025codeswitchingincontextlearningcrosslingual}, which can boost reasoning in low-resource settings. Conversely, \citet{tu-etal-2025-blessing} find that target-only or single high-resource source languages often outperform mixed prompts, suggesting that consistency within the demonstration language may be beneficial. 

Research into demonstration selection further highlights the importance of retrieval quality. While alignment-based methods \citep{tanwar-etal-2023-multilingual} and multilingual retrievers \citep{lin-etal-2025-xampler} improve results, these studies often restrict sources to English or leave selection unsystematic \citep{cahyawijaya-etal-2024-llms}. Finally, while linguistic similarity has been proposed as a selection predictor \citep{kaneko2025balancedmultifactorincontextlearning}, its effectiveness in modern decoder-only LLMs remains an open question.

\section{Methodology}

\subsection{Tasks}
To evaluate cross-lingual ICL, we use seven multilingual benchmarks (Table \ref{tab:multilingual_datasets}). Unlike prior studies that focus on English or target-only demonstration samples, we evaluate all source-target language combinations.

To manage the quadratic increase in evaluation runs, we selected 18 typologically diverse languages with maximal benchmark overlap. We further limited each benchmark to 1,000 test instances (or the full set, if smaller). Since all benchmarks are fully parallel, using aligned test and in-context examples ensures content consistency across language pairs and isolates cross-lingual transfer dynamics from content-related confounds.

\subsection{Models}
We conduct experiments with six LLMs, focusing on relatively small models (4B parameters or fewer) to keep the evaluation computationally manageable. To assess robustness with respect to model stochasticity, we repeat all experiments for the three smallest models using two different random seeds, yielding distinct sets of demonstration examples. We then compute the correlation between performance scores across language pairs for the two runs. All correlations are $\geq 0.9$ across all tasks and models, indicating high stability. Based on this result, we perform single runs for the two larger models without compromising the reliability of our findings.

More specifically, we use \textbf{Llama 3.2 (1B \& 3B)}, \textbf{Gemma 3 (1B \& 4B)} \citep{gemmateam2025gemma3technicalreport}, \textbf{Qwen3 (1.7B)} \citep{yang2025qwen3technicalreport} and \textbf{Phi-4-Mini} \citep{microsoft2025phi4minitechnicalreportcompact}.

Following \citet{zhang-etal-2024-impact}, who report that adding more than 2–4 demonstrations yields minimal further gains in multilingual scenarios, we adopt a 4-shot configuration for all experiments\footnote{Prompts used in the experiments are provided in Appendix~\ref{app:prompts}.}.

\section{Results} \label{sec:results_task_specific_experiments}

\begin{figure*}[!ht]
    \centering
    \includegraphics[width=0.98\linewidth]{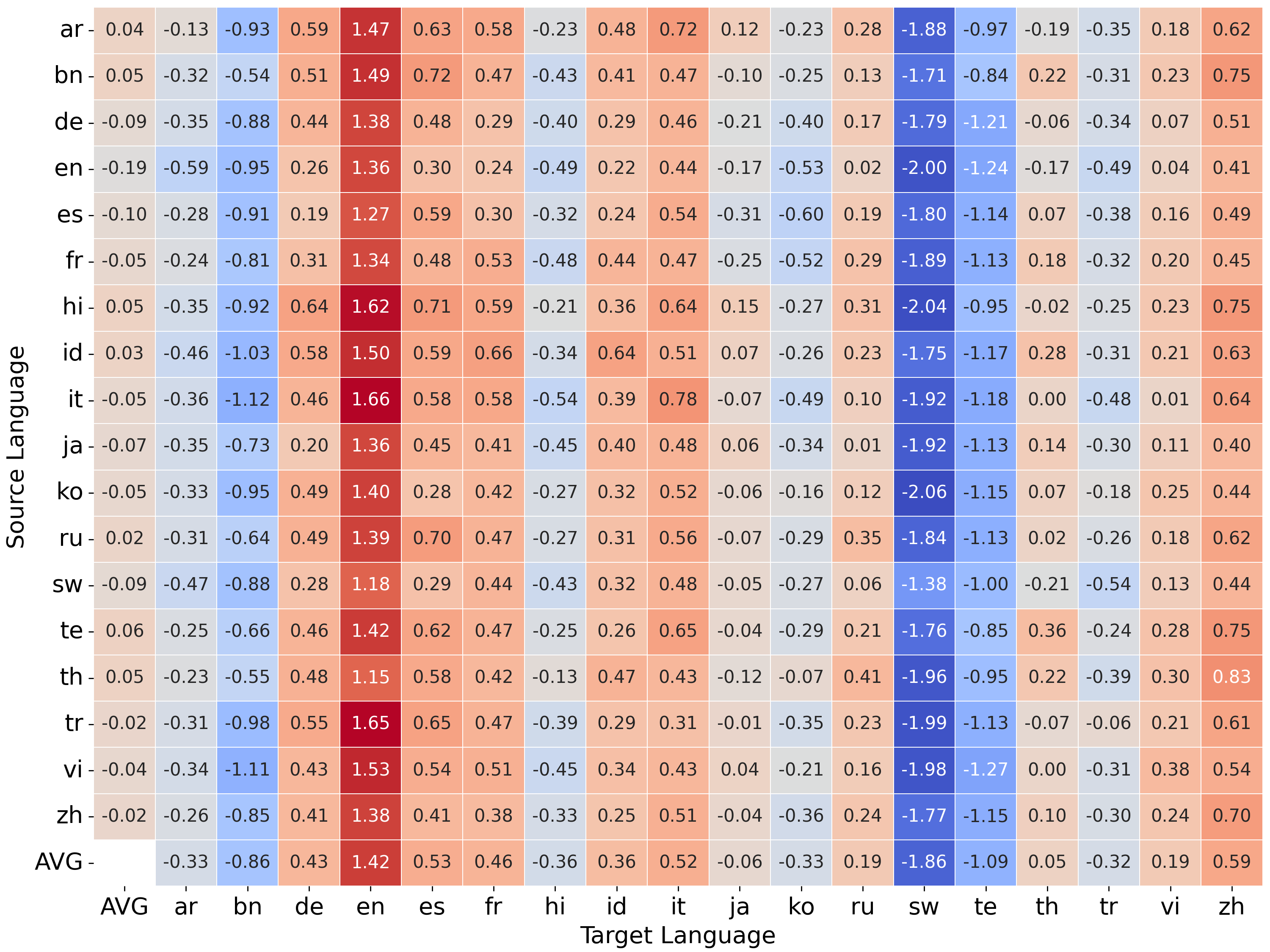}
    \caption{Cross-lingual transfer matrix showing the mean z-score of model performance across language pairs. Rows represent source languages and columns represent target languages. Scores are normalized per task and model.}
    \label{fig:overall_transfer_results}
\end{figure*}

Figure \ref{fig:overall_transfer_results} presents the average transfer performance across all evaluated language pairs, showing mean z-scores computed over seven tasks and six models.\footnote{Z-scores are computed separately for each model–task combination before averaging.} From these results, we identify different high-level observations:
\begin{itemize}
    \item \textbf{Persistent performance disparities across target languages.} As expected, English leads by a wide margin, followed by other high-resource languages such as Spanish, Italian, Indonesian, and German, while lower-resource languages like Swahili, Telugu, and Bengali show substantially lower scores.
    \item \textbf{The target language is not always its own best source}, contrary to a common assumption. In fact, across all experiments, the target language is the best source in only about $24\%$ of cases.
\end{itemize}

\begin{figure}[htbp!]
    \centering

    \begin{subfigure}[b]{0.98\linewidth}
        \centering
        \includegraphics[width=0.97\linewidth]{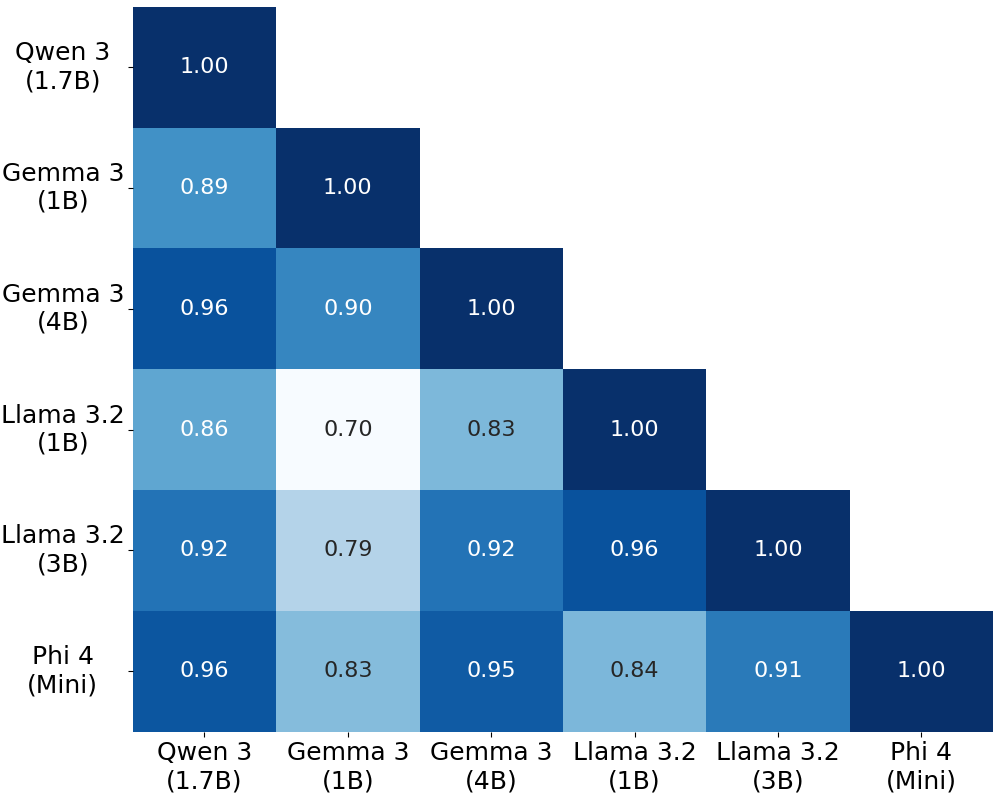}
        \caption{Across models.}
        \label{fig:correlation_between_models}
    \end{subfigure}
    
    \vspace{0.5cm}
    
    \begin{subfigure}[b]{0.98\linewidth}
        \centering
        \includegraphics[width=0.97\linewidth]{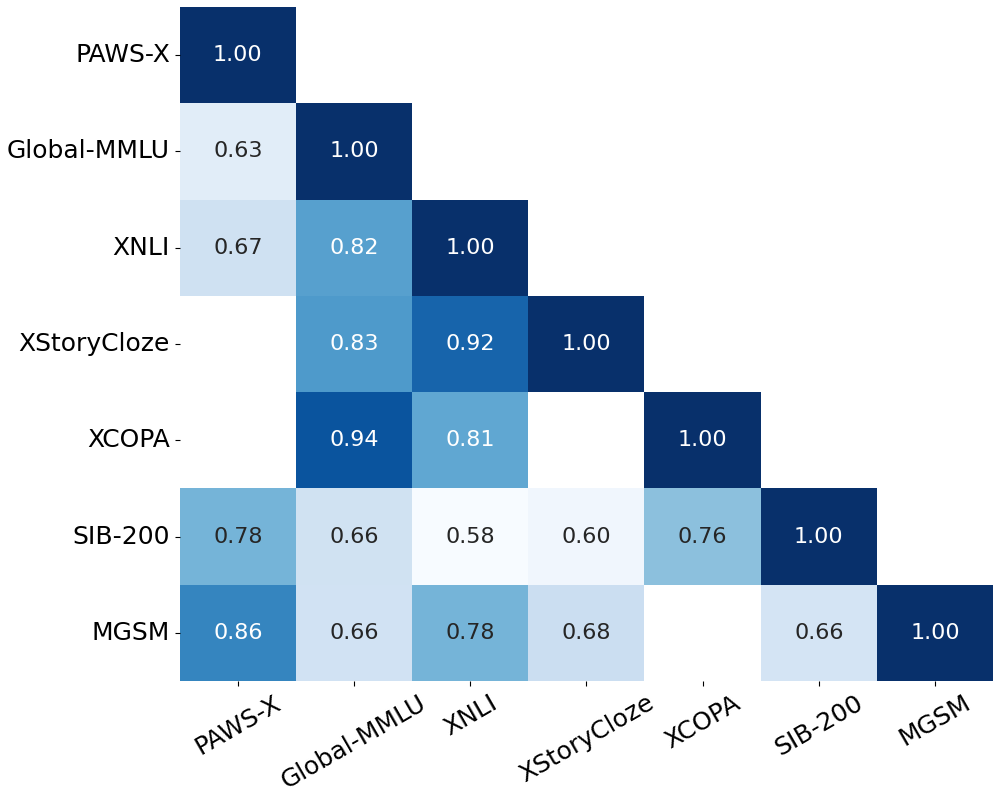}
        \caption{Across tasks.}
        \label{fig:correlation_between_tasks}
    \end{subfigure}
    
    \caption{Correlation matrix of transfer patterns. Pearson correlations are computed over all language-pair transfer scores.}
\end{figure}

\begin{itemize}
    \item \textbf{Models show highly stable transfer patterns.} Similarities are strongest within model families and among larger models (Figure~\ref{fig:correlation_between_models}).
    \item \textbf{Task-level transfer similarities are moderate to high.} Tasks with related reasoning or linguistic structure show more aligned transfer patterns (Figure~\ref{fig:correlation_between_tasks}).
\end{itemize}

\vspace{0.5cm}

A deeper statistical analysis (\S\ref{sec:donor_recipient_correlation} - \S \ref{sec:script_resource}) uncovers several critical findings:
\begin{enumerate}
    \item \textbf{Strong target languages tend to be weak source languages (\S \ref{sec:donor_recipient_correlation}).} English, Spanish, German, and Italian perform well as targets but are among the weakest sources, whereas Thai, Telugu, and Bengali show the opposite pattern.
    \item \textbf{Linguistic similarity between source and target does not predict transfer effectiveness (\S \ref{sec:ling_dist_correlation}).} Unlike prior work on fine-tuning–based transfer, we observe no such relationship in the ICL setting.
    \item \textbf{Cross-lingual alignment within the model is a stronger indicator of transfer success (\S \ref{sec:CL_alignment_correlation}).} This alignment explains far more variance in performance than surface linguistic properties.
    \item \textbf{Low-resource, non-Latin-script languages make the most effective sources, whereas high-resource Latin-script languages are the least effective (\S \ref{sec:script_resource}).} While script and resource level each have independent effects, their interaction amplifies the pattern.
\end{enumerate}

\subsection{Correlation between Donor and Recipient Capabilities} \label{sec:donor_recipient_correlation}
Inspired by the approaches of \citet{malkin-etal-2022-balanced} and \citet{dymkiewicz2025donorsrecipientsasymmetrictransfer}, we analyze donor and recipient relationships across languages. To compute these scores, we follow the 
method introduced by \citet{malkin-etal-2022-balanced}: for each language $\mathcal{L}$, we define its donor score $\mathcal{D}(\mathcal{L})$ as the average performance it provides to all other target languages (excluding cases where $\mathcal{L}$ is the target). Conversely, the recipient score $\mathcal{R}(\mathcal{L})$ is defined as the average performance that $\mathcal{L}$ receives from all other source languages (excluding cases where $\mathcal{L}$ is the source). 

\begin{figure}[H]
    \centering
    \includegraphics[width=0.98\linewidth]{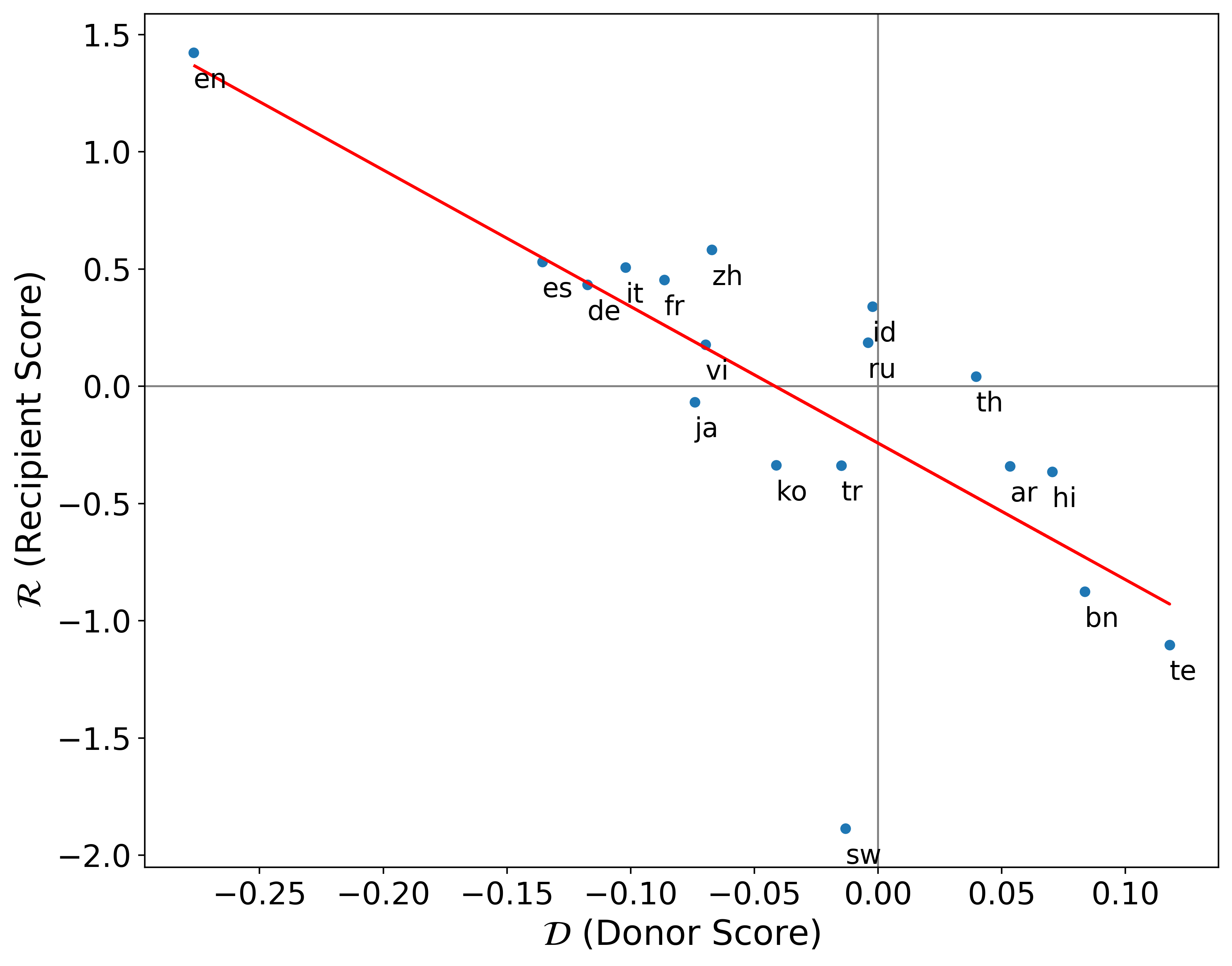}
    \caption{Scatter plot of donor and recipient capabilities; the red line indicates the best-fit regression capturing their inverse relationship.}
    \label{fig:donor_recipient}
\end{figure}

Figure~\ref{fig:donor_recipient} shows that donor and recipient capabilities are strongly negatively related, and we observe a Pearson correlation of $-0.932$ between source-to-target and target-to-source transfer across all models and tasks.

\subsection{Correlation with Linguistic Similarity} \label{sec:ling_dist_correlation}
We examine whether the well-established relationship between linguistic similarity and transfer performance in fine-tuning scenarios also holds for prompting-based approaches. Using URIEL and lang2vec representations \citep{littell-etal-2017-uriel}, we compute four categories of linguistic similarity (syntactic, genetic, phonological, and featural) and correlate them with performance for each task and model pair individually.

\begin{figure}[H]
    \centering
    \includegraphics[width=0.98\linewidth]{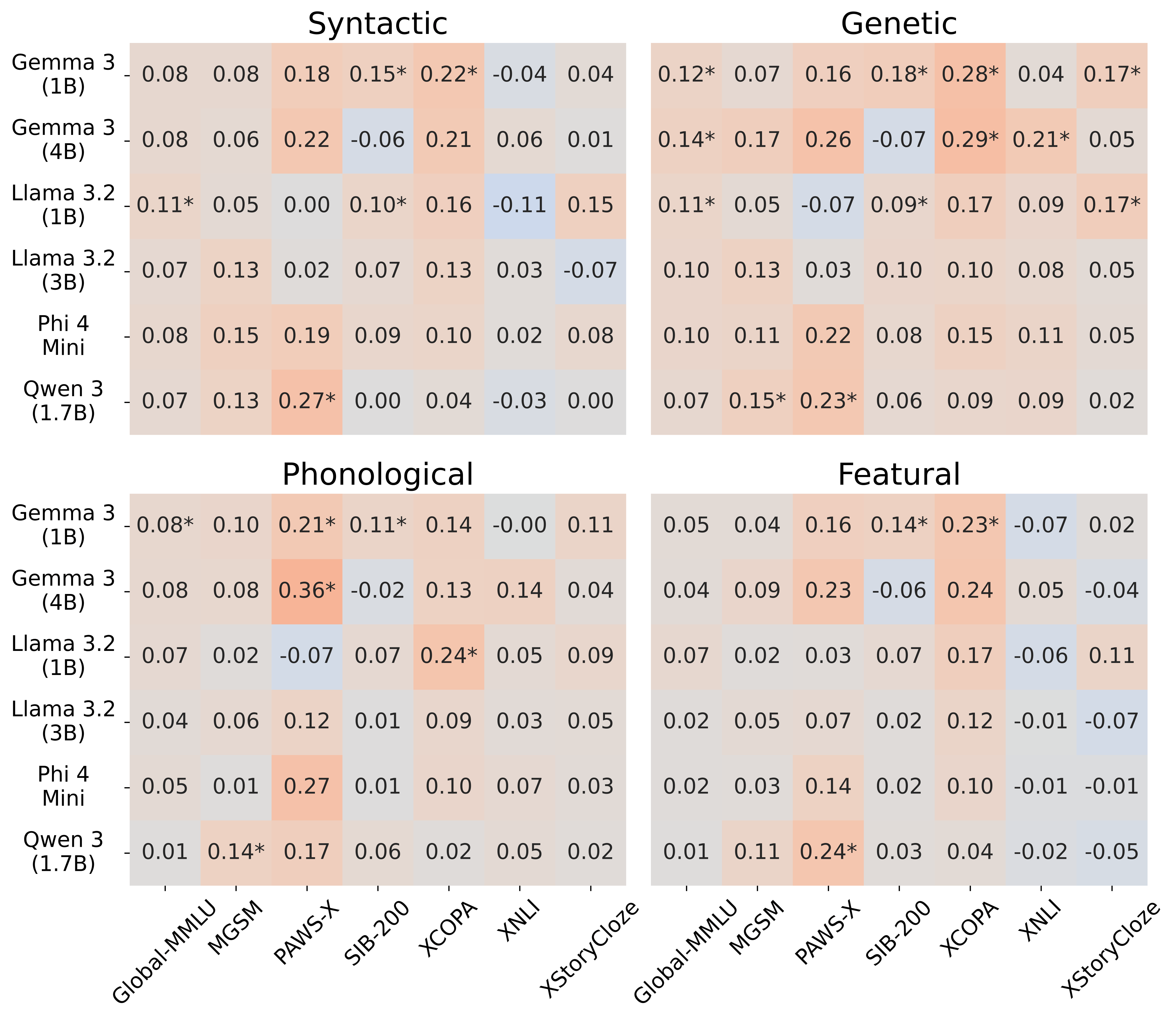}
    \caption{The four heatmaps illustrate the \textbf{correlation between linguistic similarity} (syntactic, genetic, phonological, and featural) of source–target language pairs \textbf{and cross-lingual transfer performance}. Each heatmap corresponds to one \textit{lang2vec} dimension and reports the Pearson correlation coefficient for every benchmark (column) and model (row) combination. An asterisk (*) indicates statistical significance at $p < 0.05$.}
    \label{fig:LingDist_correlations}
\end{figure}

As shown in Figure \ref{fig:LingDist_correlations}, only a small subset of task and model combinations exhibits low to moderate correlations across these features, and even these effects are inconsistent. For the vast majority of settings, we observe no strong correlation between linguistic similarity and transfer success. These findings suggest that, unlike fine-tuned models, in-context learning does not systematically rely on linguistic proximity, which highlights a fundamental difference in how cross-lingual generalization emerges in the two paradigms.

\subsection{Correlation with Cross-Lingual Alignment} \label{sec:CL_alignment_correlation}

\begin{figure}[H]
    \centering
    \includegraphics[width=0.98\linewidth]{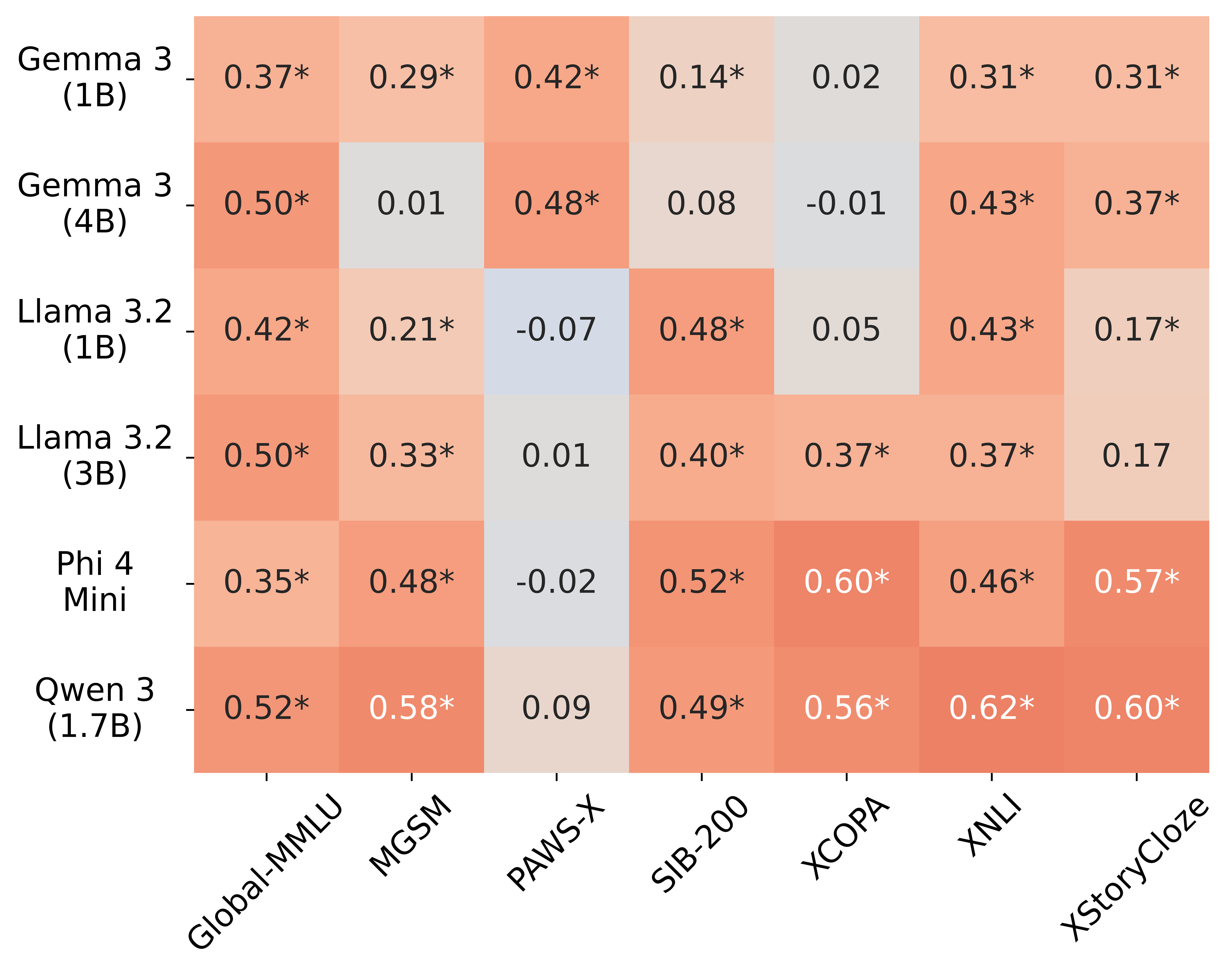}
    \caption{This heatmap shows how strongly a model's internal alignment (last layer) between a given source–target language pair correlates with its transfer performance for that same pair. Correlation values are shown for each benchmark (column) and each model (row) combination.}
    \label{fig:CKA_correlations}
\end{figure}

Beyond surface-level linguistic features, we also analyze whether transfer performance is related to the degree of cross-lingual alignment between the source and target languages within the model’s internal representations. Although cross-lingual alignment is often correlated with underlying linguistic similarity, it captures a fundamentally different notion: it reflects how closely languages are positioned in the model’s learned representation space, which may diverge from typological similarity due to factors such as training data composition, tokenization, or implicit model biases. To measure this, we use the devtest portion of FLORES-200 \citep{nllbteam2022languageleftbehindscaling}, which provides aligned sentences across 200 languages, and for each model we compute the similarity of mean-pooled last-layer hidden states across languages using the Centered Kernel Alignment (CKA) metric \citep{kornblith2019similarity}. We find that these correlations between cross-lingual alignment and transfer performance are substantially stronger than those obtained from linguistic similarity features, although the strength of the effect varies depending on the task and model (Figure \ref{fig:CKA_correlations}). This indicates that, in the context of in-context learning, models rely far more on the cross-lingual structure they have learned during pre-training and the resulting proximity of languages in their representation space than on any surface-level linguistic similarities.

\subsection{Effect of Script and Resource Availability on Source Language Performance}
\label{sec:script_resource}

To assess whether (i) the writing system and (ii) the resource status of the source language influence cross-lingual transfer performance, we perform a set of inferential statistical analyses.

First, we define a binary variable indicating whether the source language uses a Latin-based script. Resource status is likewise modeled as a binary variable, distinguishing between higher- and lower-resource languages based on their relative representation in the Common Crawl corpus\footnote{Based on CC-MAIN-2025-38, retrieved from \url{https://commoncrawl.github.io/cc-crawl-statistics/plots/languages}.}, which we treat as a proxy for the relative availability of a language in typical large-scale pre-training data.

We first fit a two-way ANOVA model with transfer performance as the dependent variable, including script type (Latin vs.~non-Latin), binary resource status (high vs.~low), and their interaction as predictors. The analysis reveals significant main effects of both \textit{Script} and \textit{Resource Status}. The interaction between script type and resource status is not significant, indicating that the effects of script and resource status on transfer performance are additive rather than interactive (Table \ref{tab:anova})\footnote{Assumptions were evaluated using residual diagnostics. Q–Q plots indicated approximate normality with mild tail deviations. Homogeneity of variances was assessed using Levene's test and was satisfied (p = .59).}.

\begin{table}[h!]
\centering
\small
\begin{tabular}{lrr}
\toprule
\textbf{Model Term} & \textbf{F} & \textbf{p-value} \\
\midrule
Script (Latin vs.\ Non-Latin) & 15.29 & 9.3e-05 \\
Resource Status (High vs.\ Low) & 17.86 & 2.4e-05 \\
Script $\times$ Resource Status & 0.09 & 0.76 \\
\bottomrule
\end{tabular}
\vspace{2mm}
\caption{\textbf{Simplified ANOVA results (script type $\times$ resource level interaction model)} showing F-statistics and p-values for script type, resource level, and their interaction.}
\label{tab:anova}
\end{table}

Second, to facilitate pairwise comparisons, we conduct post-hoc Tukey HSD tests to assess pairwise differences in transfer performance between all group combinations (Tables \ref{tab:tukey_resource}, \ref{tab:tukey_script} \& \ref{tab:tukey_full}).

Averaged across script types, low-resource source languages significantly outperform high-resource source languages (mean difference = 0.1187, $p < .001$), and averaged across resource levels, non-Latin source languages significantly outperform Latin-script source languages (mean difference = 0.1133, $p < .001$). 
When examining combined script–resource groups, both script type and resource level yield significant differences within each other’s levels. However, the comparison between high-resource non-Latin and low-resource Latin source languages is not significant, indicating that advantages associated with script type and resource availability can partially offset one another.

\begin{table}[!ht]
\centering
\small
\begin{tabular}{lrrrr}
\toprule
\textbf{Comparison} & \textbf{meandiff} & \textbf{p-adj} & \textbf{lower} & \textbf{upper} \\
\midrule
High → Low & 0.1187 & 0.0000 & 0.0767 & 0.1607 \\
\bottomrule
\end{tabular}
\vspace{2mm}
\caption{Tukey HSD contrast for resource level (High vs.\ Low). Low-resource languages show significantly higher transfer performance.}
\label{tab:tukey_resource}
\end{table}

\begin{table}[!ht]
\centering
\small
\begin{tabular}{lrrrr}
\toprule
\textbf{Comparison} & \textbf{meandiff} & \textbf{p-adj} & \textbf{lower} & \textbf{upper} \\
\midrule
NL → L & -0.1133 & 0.0000 & -0.1552 & -0.0714 \\
\bottomrule
\end{tabular}
\vspace{2mm}
\caption{Tukey HSD contrast for script type (Non-Latin [NL] vs.\ Latin [L]). The negative value indicates higher performance for NL languages.}
\label{tab:tukey_script}
\end{table}

\begin{table}[!ht]
\centering
\small
\begin{tabular}{lrrrr}
\toprule
\textbf{Comparison} & \textbf{meandiff} & \textbf{p-adj} \\
\midrule
High\_NL → High\_L & -0.0807 & 0.0394 \\
High\_NL → Low\_NL & 0.1010 & 0.0075 \\
High\_NL → Low\_L   & 0.0068 & 0.9976 \\[2pt]

High\_L → Low\_NL  & 0.1817 & 0.0000 \\
High\_L → Low\_L    & 0.0875 & 0.0284 \\[2pt]

Low\_NL → Low\_L   & -0.0942 & 0.0206 \\
\bottomrule
\end{tabular}
\vspace{2mm}
\caption{Simplified Tukey HSD post-hoc comparison across all Script (Non-Latin [NL] \& Latin [L]) and Resource (High \& Low) combinations, showing only mean differences and adjusted p-values.}
\label{tab:tukey_full}
\end{table}


\begin{figure*}[htbp!]
    \centering
    \begin{subfigure}[b]{0.48\linewidth}
        \centering
        \includegraphics[width=\linewidth]{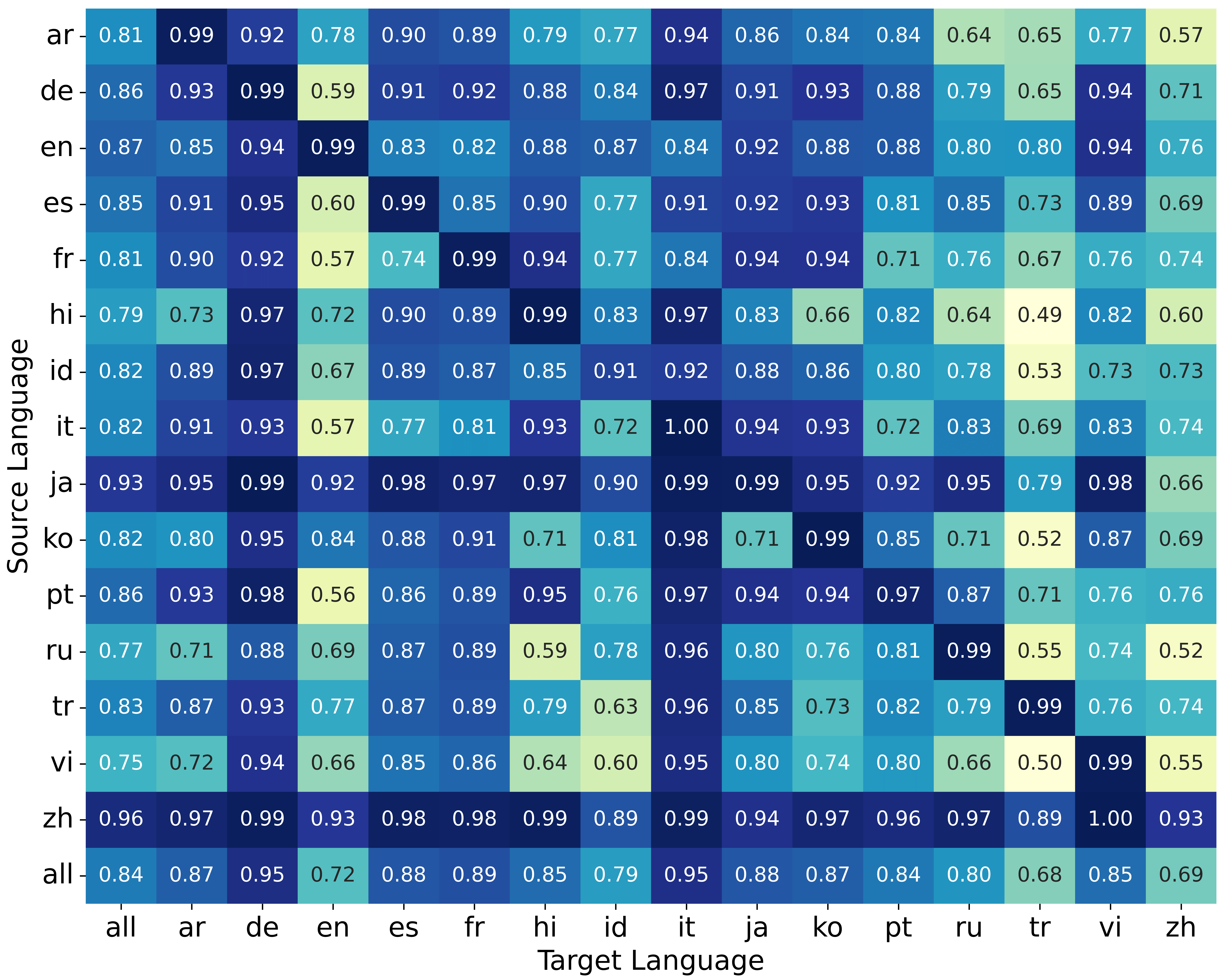}
        \caption{Monolingual setup.}
        \label{fig:results_lang_confusion_monolingual}
    \end{subfigure}
    \hfill 
    \begin{subfigure}[b]{0.48\linewidth}
        \centering
        \includegraphics[width=\linewidth]{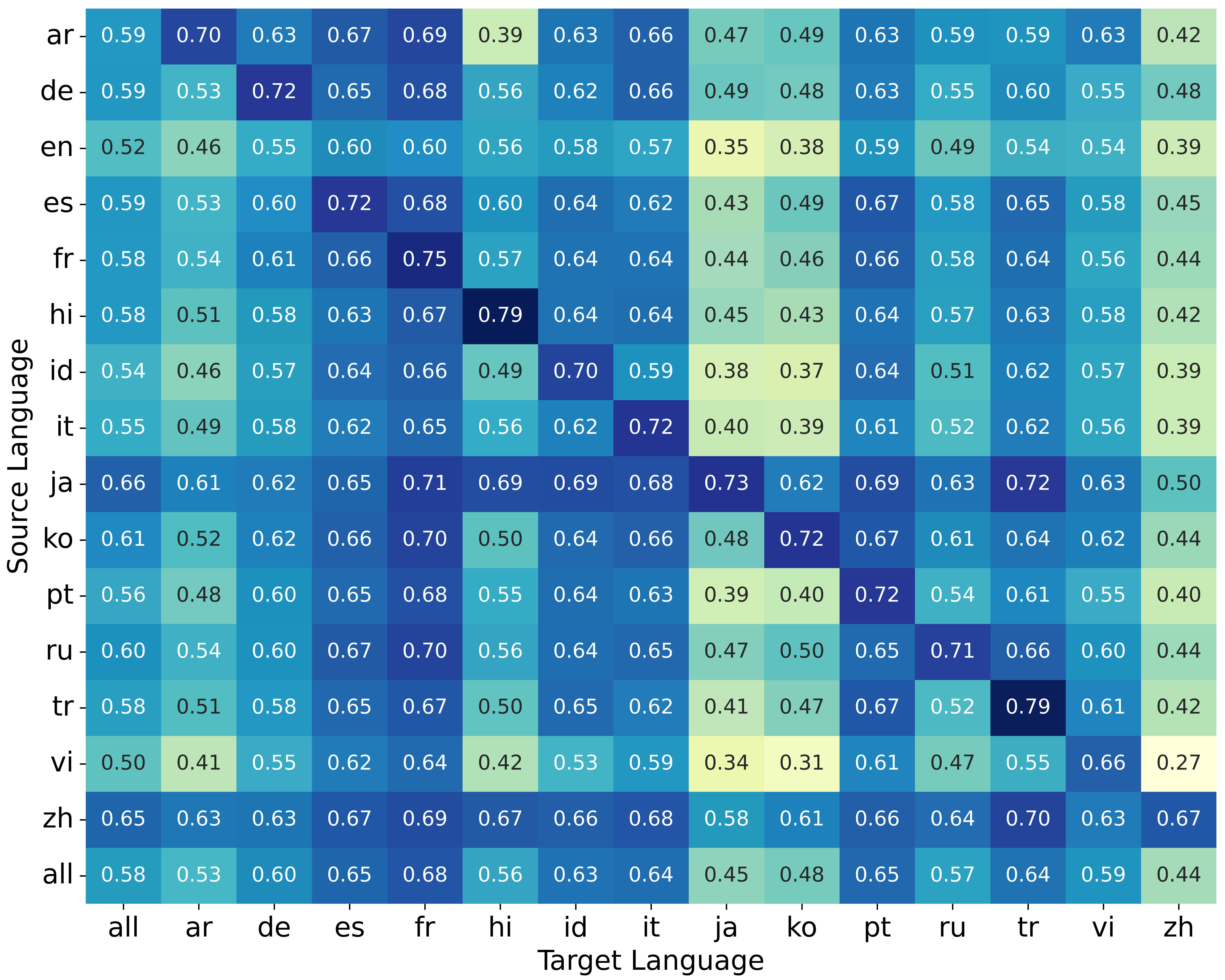}
        \caption{Cross-lingual setup.}
        \label{fig:results_lang_confusion_crosslingual}
    \end{subfigure}

    \caption{Average Line-level Pass Rates (LPR) across models for different source-target language pairs.}
\end{figure*}

\section{Language Confusion in Cross-Lingual In-Context Learning}
To broaden our analysis beyond classification tasks, we additionally investigate whether source-language variation in ICL produces comparable effects in generative settings.

In our cross-lingual setup, we do not prioritize task-specific performance. Instead, we shift our attention to diagnosing a fundamental obstacle in generative cross-lingual evaluation: \textit{language confusion}. Language confusion occurs when a model’s responses are produced in a language different from the one requested, such as defaulting to English rather than the target language. When a model fails to generate output in the intended language $\mathcal{L}$, we cannot meaningfully assess cross-lingual transfer quality because no valid output in $\mathcal{L}$ exists to evaluate. In real applications, this failure means the user’s request is not truly addressed, as the response may be partially or entirely incomprehensible. For these reasons, language confusion is arguably a more fundamental metric here than task performance itself.

We therefore evaluate models using the \textit{Language Confusion Benchmark} \citep{marchisio-etal-2024-understanding}, which measures a model's ability to produce responses in the specified language. The benchmark spans 15 typologically diverse languages that fully overlap with the language set used in Section \ref{sec:results_task_specific_experiments}. It includes two settings: (1) \textbf{monolingual}, where both the prompt and expected response are in the target language, and (2) \textbf{cross-lingual}, where the prompt is given in English but explicitly instructs the model to answer in the target language. Consistent with earlier experiments, we use a 4-shot ICL setup with demonstration examples drawn from the parallel Bactrian-X instruction-tuning dataset \citep{li2023bactrianxmultilingualreplicableinstructionfollowing}, which covers all languages present in the benchmark.

As a metric, we adopt the \textit{Line-level Pass Rate} (\textbf{LPR}) introduced by \citet{marchisio-etal-2024-understanding}. LPR is defined as the proportion of model responses where every line is correctly identified as being in the user's intended language\footnote{Following \citet{marchisio-etal-2024-understanding}, we perform line-level Language Identification (LID) using fastText \citep{joulin2016bag} and restrict its use to sequences with more than four words to ensure LID reliability.}.

\subsection{Results}
We provide the average LPR across all models for each source-target pair in Figures \ref{fig:results_lang_confusion_monolingual} and \ref{fig:results_lang_confusion_crosslingual} for the monolingual and cross-lingual settings respectively.

\begin{figure}[H]
    \centering
    \includegraphics[width=0.99\linewidth]{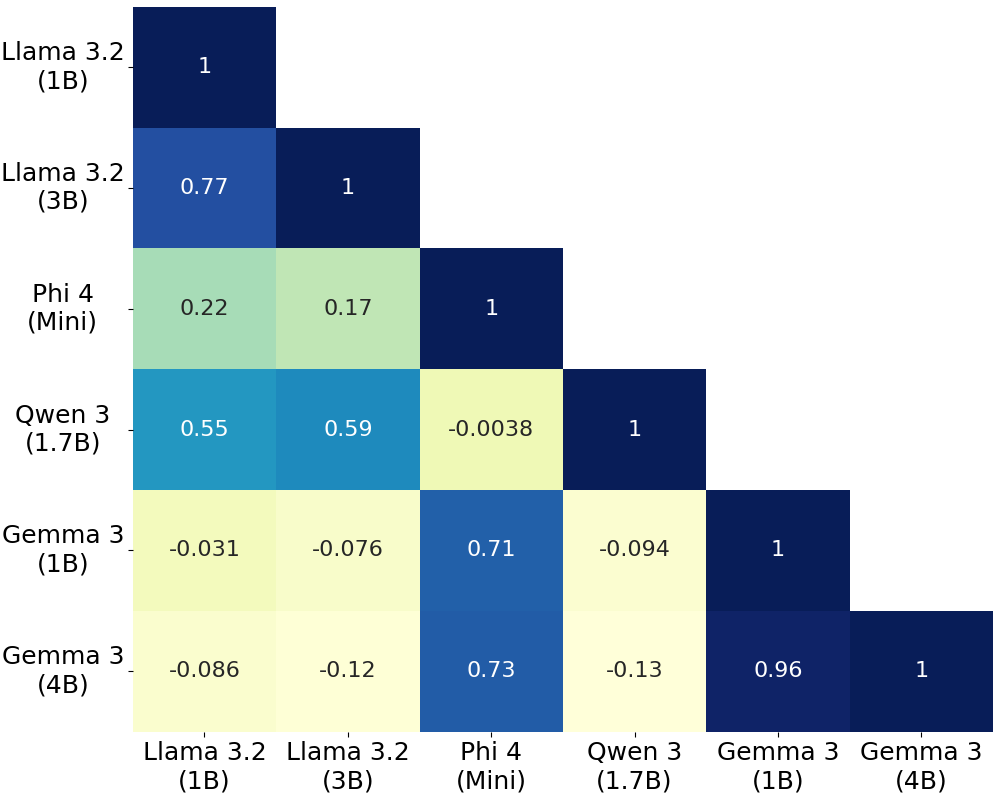}
    \caption{Correlation matrix of language confusion patterns across models. Pearson correlations are computed over all language-pair transfer scores.}
    \label{fig:lang_confusion_model_corr_heatmap}
\end{figure}

\paragraph{Overall patterns of language confusion.} \ \\
Unlike the consistent patterns observed in task-specific classification, language confusion shows substantially less consistency across models. As expected, confusion patterns are most similar within model families (Gemma 3 and Llama 3.2), with moderate correlations also observed between Qwen 3 and Llama 3.2, and between Phi 4 and Gemma 3 (Figure~\ref{fig:lang_confusion_model_corr_heatmap}).

\paragraph{Relationship to task transfer performance.} \ \\
We further compute correlations between task-level transfer performance and language confusion. Overall, we find that transfer performance for some tasks correlates moderately with language confusion in the cross-lingual setting, while other tasks show moderate correlations in the monolingual setting. However, these relationships are not consistent across tasks or models, indicating that \textbf{language confusion is not directly linked to unequal transfer performance, but instead represents a distinct challenge}.

Additionally, across all models and both settings, we find a Pearson correlation of 0.6342 between a language’s tendency to induce confusion when used as a source language and the confusion it experiences as a target language. This source–target symmetry mirrors the correlation previously observed between donor and recipient capabilities of a language in the task transfer experiments (\S \ref{sec:donor_recipient_correlation}), albeit with a somewhat weaker magnitude for language confusion.

\paragraph{Linguistic correlates of language confusion.} \ \\
We investigate correlations between language confusion and two factors: linguistic similarity and the cross-lingual alignment of source and target language embedding spaces. For both the monolingual and cross-lingual settings, no strong correlations were observed for any model tested\footnote{All calculated Pearson correlation values were below 0.3.}. Similarly, we found no strong tendency regarding the correlation between the resource level of the source or target language and the resulting language confusion.

However, t-tests comparing language performance based on writing system yielded significant results ($\alpha=0.05$). Specifically, we found that non-Latin script languages suffer more from language confusion as a target language, but they create less language confusion in other target languages when used as the source language, compared to Latin script languages.

\section{Discussion}
Our findings suggest that ICL leverages mechanisms distinct from fine-tuning during cross-lingual transfer. The lack of correlation with linguistic similarity indicates that ICL relies less on structural overlap and more on representational properties learned during pretraining. While linguistic similarity guides parameter updates, it is less relevant when models must infer tasks from contextual examples at inference time.

One possible explanation is that source languages differ in the informational constraints they impose during in-context learning. High-resource languages such as English may be highly entropic, activating many overlapping pretraining patterns and encouraging reliance on broad, language-specific heuristics rather than task structure. In contrast, low-resource or typologically atypical languages may act as implicit regularizers, reducing spurious associations and forcing the model to focus more strongly on the abstract input–output mapping illustrated by the examples.

The strong effectiveness of low-resource, non-Latin script languages as source languages raises practical and theoretical questions. While this finding suggests that practitioners could improve cross-lingual ICL performance by selecting such languages as sources, it remains unclear whether this strategy is robust or future-proof. As more data becomes available for currently low-resource languages, their role within multilingual models may change, potentially diminishing the very properties that make them effective sources today. Moreover, it is unknown whether there exists a lower bound of resource availability beyond which in-context examples cease to be helpful, because the model no longer reliably understands the language itself. This points to an unresolved trade-off between linguistic distinctiveness and model familiarity.

Finally, our results underscore that source and target languages play fundamentally different roles in cross-lingual ICL, and that strong performance as a target language does not imply effectiveness as a source language. This asymmetry suggests that current evaluation practices, often focused solely on target-side performance, may overlook critical aspects of source-language selection. Future work should aim to better characterize the properties that make a language a good source in ICL, and to disentangle whether these effects stem from script, data imbalance, pretraining dynamics, or more general regularization-like effects induced by atypical language distributions.
\section{Conclusion}
In this work, we showed that cross-lingual ICL exhibits behaviors that differ markedly from established patterns in fine-tuning-based cross-lingual transfer. In particular, source language effectiveness is not driven by linguistic similarity, and languages that perform poorly as targets are the most effective sources. These findings challenge common assumptions about source language selection and suggest that cross-lingual ICL relies on mechanisms that are still poorly understood. Beyond their immediate practical implications, our results point to the need for a deeper theoretical account of how multilingual models exploit in-context examples across languages, especially in settings involving low-resource and typologically distant languages.

\section*{Limitations}
Our analysis is conducted on a carefully selected set of 18 languages that balances typological diversity and benchmark overlap. While this enables controlled and systematic comparisons, extending the study to a broader range of languages, particularly those with extremely limited model support, would further strengthen the generality of our conclusions.

For computational tractability, we focus on relatively small language models ($\leq$ 4B parameters). The consistency of transfer patterns across these models suggests that our findings capture stable trends, though evaluating larger models remains an important direction for future work.

\section*{Ethics Statement}
Although we find that lower-resource languages can be effective source languages in cross-lingual ICL, this should \textbf{not} be interpreted as a recommendation to deliberately reduce the presence of certain languages during pretraining for the benefit of other target languages. Such practices would risk reinforcing existing inequities in multilingual NLP. Our findings instead reflect properties of current models and training distributions.

\bibliography{latex/custom}

\appendix

\section{Prompts} \label{app:prompts}

\begin{figure}[H]
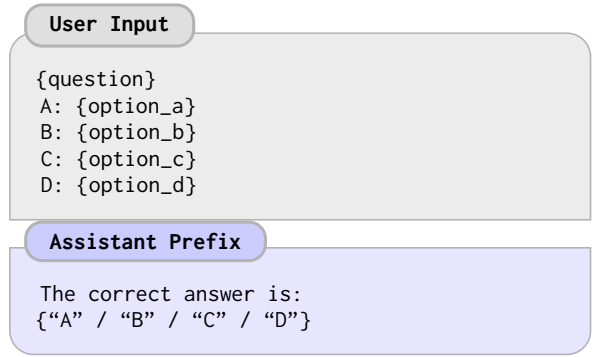

    \centering
    \begin{userinputbox}
    \{question\} \\
    A: \{option\_a\} \\
    B: \{option\_b\} \\
    C: \{option\_c\} \\
    D: \{option\_d\}
    \end{userinputbox}
    \begin{assistantprefixbox}
    The correct answer is: \\ \{``A'' / ``B'' / ``C'' / ``D''\}
    \end{assistantprefixbox}
    \caption{Prompt template used for Global-MMLU.}
    \label{fig:prompt_MMLU}
\end{figure}

\begin{figure}[H]
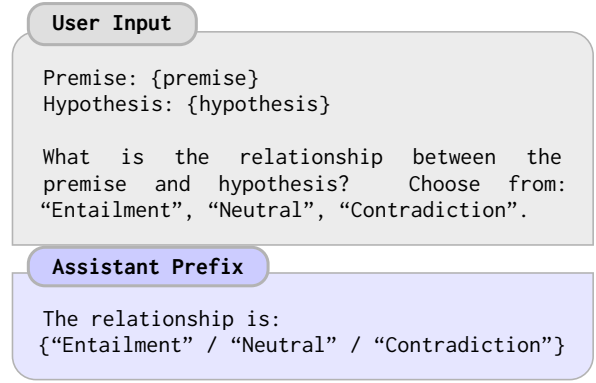

    \centering
    \begin{userinputbox}
    Premise: \{premise\} \\
    Hypothesis: \{hypothesis\} \\ \\
    What is the relationship between the premise and hypothesis? Choose from: ``Entailment'', ``Neutral'', ``Contradiction''.
    \end{userinputbox}
    \begin{assistantprefixbox}
    The relationship is: \\ \{``Entailment'' / ``Neutral'' / ``Contradiction''\}
    \end{assistantprefixbox}
    \caption{Prompt template used for XNLI.}
    \label{fig:prompt_XNLI}
\end{figure}

\begin{figure}[H]
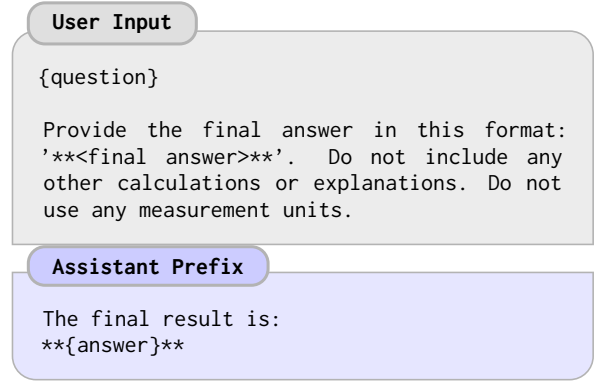

    \centering
    \begin{userinputbox}
    \{question\} \\ \\
    Provide the final answer in this format: '**<final answer>**'. Do not include any other calculations or explanations. Do not use any measurement units.
    \end{userinputbox}
    \begin{assistantprefixbox}
    The final result is: \\ **\{answer\}**
    \end{assistantprefixbox}
    \caption{Prompt template used for MGSM.}
    \label{fig:prompt_MGSM}
\end{figure}

\begin{figure}[H]
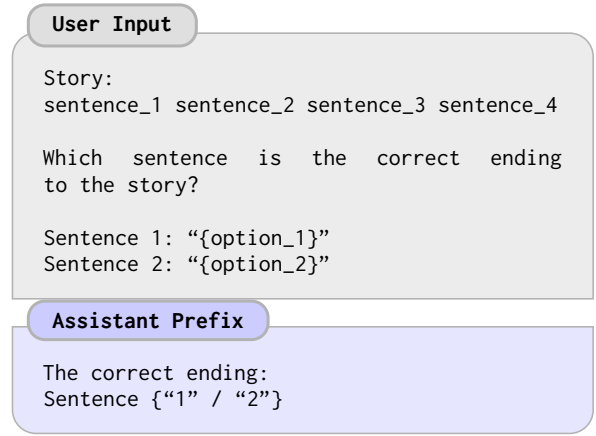

    \centering
    \begin{userinputbox}
    Story: \\
    {sentence\_1} {sentence\_2} {sentence\_3} {sentence\_4} \\ \\
    Which sentence is the correct ending to the story? \\ \\
    Sentence 1: ``\{option\_1\}'' \\
    Sentence 2: ``\{option\_2\}''
    \end{userinputbox}
    \begin{assistantprefixbox}
    The correct ending: \\ Sentence \{``1'' / ``2''\}
    \end{assistantprefixbox}
    \caption{Prompt template used for XStoryCloze.}
    \label{fig:prompt_XStoryCloze}
\end{figure}

\begin{figure}[H]
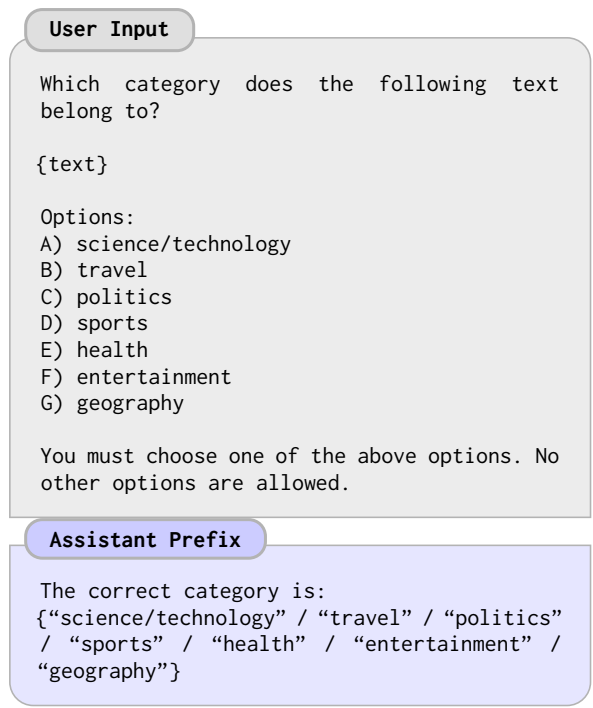

    \centering
    \begin{userinputbox}
    Which category does the following text belong to? \\ \\
    \{text\} \\ \\
    Options: \\
    A) science/technology \\
    B) travel \\
    C) politics \\
    D) sports \\
    E) health \\
    F) entertainment \\
    G) geography \\ \\
    You must choose one of the above options. No other options are allowed.
    \end{userinputbox}
    \begin{assistantprefixbox}
    The correct category is: \\ \{``science/technology'' / ``travel'' / ``politics'' / ``sports'' / ``health'' / ``entertainment'' / ``geography''\}
    \end{assistantprefixbox}
    \caption{Prompt template used for SIB-200.}
    \label{fig:prompt_SIB200}
\end{figure}

\begin{figure}[H]
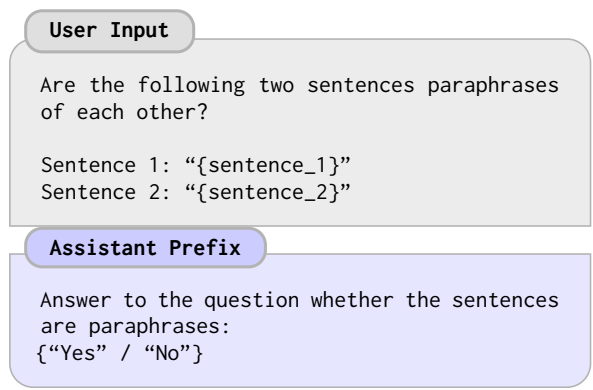

    \centering
    \begin{userinputbox}
    Are the following two sentences paraphrases of each other? \\ \\
    Sentence 1: ``\{sentence\_1\}'' \\
    Sentence 2: ``\{sentence\_2\}''
    \end{userinputbox}
    \begin{assistantprefixbox}
    Answer to the question whether the sentences are paraphrases: \\ \{``Yes'' / ``No''\}
    \end{assistantprefixbox}
    \caption{Prompt template used for PAWS-X.}
    \label{fig:prompt_PAWSX}
\end{figure}

\begin{figure}[H]
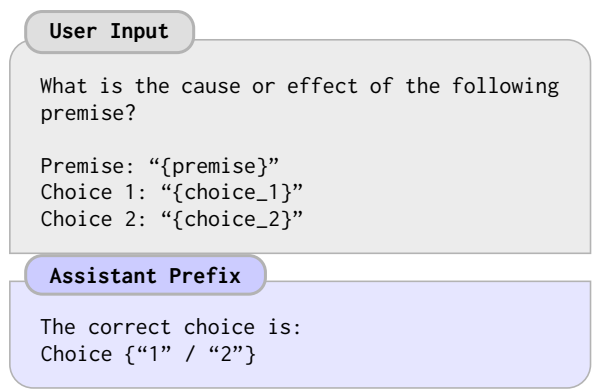

    \centering
    \begin{userinputbox}
    What is the cause or effect of the following premise? \\ \\
    Premise: ``\{premise\}'' \\
    Choice 1: ``\{choice\_1\}'' \\
    Choice 2: ``\{choice\_2\}''
    \end{userinputbox}
    \begin{assistantprefixbox}
    The correct choice is: \\
    Choice \{``1'' / ``2''\}
    \end{assistantprefixbox}
    \caption{Prompt template used for XCOPA.}
    \label{fig:prompt_XCOPA}
\end{figure}

The prompt templates used for the classification experiments are provided in Figures \ref{fig:prompt_MMLU}, \ref{fig:prompt_XNLI}, \ref{fig:prompt_MGSM}, \ref{fig:prompt_XStoryCloze}, \ref{fig:prompt_SIB200}, \ref{fig:prompt_PAWSX} and \ref{fig:prompt_XCOPA}.

\end{document}